\pdfoutput=1
\documentclass{article}

\PassOptionsToPackage{numbers, compress}{natbib}
\usepackage[preprint]{neurips_2022}

\usepackage[utf8]{inputenc}
\usepackage[T1]{fontenc}
\usepackage{hyperref}
\usepackage{url}
\usepackage{booktabs}
\usepackage{amsfonts}
\usepackage{nicefrac}
\usepackage{microtype}
\usepackage{xcolor}

\usepackage[all]{hypcap}
\usepackage{enumitem}
\usepackage[super]{nth}
\usepackage[color=lime,size=small]{todonotes}
\hypersetup{
  colorlinks=true,
  linkcolor=blue,
  citecolor=blue,
  filecolor=blue,
  urlcolor=blue,
}

\usepackage[section,frozencache]{minted}  
\definecolor{bg}{rgb}{0.95,0.95,0.95}
\setminted[python]{
  bgcolor=bg,
  frame=single,
  framesep=3pt,
  framerule=0pt,
  numberblanklines=false,
  python3=true,
  breaklines,
}

\newcommand{\marginnote}[1]{%
  \marginpar[\raggedleft\ttfamily\textcolor{black}{#1}]{}%
}
\newcommand{\argument}[1]{%
  \texttt{#1}\marginnote{#1}%
}

\newcommand{\torchtime}{%
  \texttt{torchtime}%
}

\title{Benchmark time series data sets for PyTorch --\\the \torchtime{} package}

\author{%
  Philip Darke \qquad Paolo Missier\thanks{Equal contribution.} \qquad Jaume Bacardit\footnotemark[1]\\
  School of Computing\\
  Newcastle University\\
  Newcastle, United Kingdom \\
  \texttt{\{p.a.darke2, paolo.missier, jaume.bacardit\}@newcastle.ac.uk}\\
}

\begin{document}
\maketitle
\reversemarginpar

\begin{abstract}
  The development of models for Electronic Health Record data is an area of active research featuring a small number of public benchmark data sets. Researchers typically write custom data processing code but this hinders reproducibility and can introduce errors. The Python package \torchtime{} provides reproducible implementations of commonly used PhysioNet and UEA \& UCR time series classification repository data sets for PyTorch. Features are provided for working with irregularly sampled and partially observed time series of unequal length. It aims to simplify access to PhysioNet data and enable fair comparisons of models in this exciting area of research.
\end{abstract}

\section{Why \torchtime{}?}

The development and benchmarking of models for time series that are irregularly sampled, partially observed and of unequal length is an area of active research. An important application is Electronic Heath Record (EHR) data which feature challenging patterns of missingness and sources of bias including informative observation \cite{Goldstein2017}. Access to EHR data is tightly controlled and there are a small number of public data sets. Each year, the major machine learning conferences include a number of papers using PhysioNet \cite{Goldberger2000} challenge and MIMIC \cite{johnson2016,johnson2020} EHR data. In addition, resources such as the UEA \& UCR time series classification repository \cite{Bagnall} provide data sets across a range of domains that are useful for demonstrating model generality and in ablation studies\footnote{For example, controlling the level of missingness in a data set.}.

Current best practice for reproducibility and the fair comparison of models \cite{heil2021reproducibility} is to make available all the data\footnote{Within reason, as medical data can often not be made available due to its sensitive nature.} and code supporting results, including any code used to prepare data for modelling. Data processing approaches vary, often for good reason. However, the repetition of writing broadly similar data classes is wasteful and can introduce errors.

The Python package \torchtime{} addresses these issues by providing implementations of commonly used data sets for PyTorch. It simplifies access to PhysioNet and UEA \& UCR repository data, removing the need for researchers to write their own data classes and enabling better research by providing reproducible implementations to level the model development and benchmarking playing field.

\subsection{Related work}

\texttt{sktime} \cite{Loning2019} prepares UEA \& UCR time series classification repository data sets for use in Python and \texttt{medical\_ts\_datasets}\footnote{\url{https://github.com/ExpectationMax/medical_ts_datasets}} \cite{horn2020} prepares PhysioNet data for TensorFlow. \torchtime{} builds on this work by providing PyTorch users with finer control of data splits, missing data simulation/imputation and the ability to append observational masks and time delta channels. The authors are unaware of a comprehensive Python package that prepares PhysioNet and other common benchmark time series classification data sets for PyTorch.

\subsection{Paper structure}

Section \ref{example} demonstrates how \torchtime{} simplifies the training of a basic RNN model on PhysioNet 2012 challenge data. Section \ref{background} provides introductory material and section \ref{using_torchtime} covers \torchtime{} usage in detail including code examples. Sections \ref{missing_data} and \ref{unequal_length} provide additional information on working with missing data and sequences of unequal length. The appendices include further detail on the implementation, the API and the PhysioNet 2012 challenge data.

\section{Introduction: \torchtime{} in action}
\label{example}

The following example fits the {GRU-Simple} model \cite{Che2018} to PhysioNet 2012 challenge data. {GRU-Simple} is a Gated Recurrent Unit (GRU) with an observational mask and the time since last observation (the ``time delta'') appended to input data.

\begin{minted}[linenos]{python}
import numpy as np
import torch
import torch.nn as nn
from sklearn.metrics import roc_auc_score
from torch.utils.data import DataLoader
from torchtime.collate import packed_sequence
from torchtime.data import PhysioNet2012

BATCH_SIZE = 128
LEARNING_RATE = 1e-3
HIDDEN_SIZE = 64
N_EPOCHS = 250
SEED = 293120

device = "cuda" if torch.cuda.is_available() else "cpu"
device = torch.device(device)
torch.manual_seed(SEED)
\end{minted}

Data is loaded using the \texttt{torchtime.data.PhysioNet2012} class. The \texttt{mask} and \texttt{delta} arguments add an observational mask and time delta. As PhysioNet 2012 challenge data are sequences of unequal length, the \texttt{torchtime.collate.pack\_padded} collate function is used to return training batches as PyTorch PackedSequence objects.

\begin{minted}[linenos,firstnumber=16]{python}
physionet2012 = PhysioNet2012(
    split="train",
    train_prop=2/3,
    impute="forward",  # use forward imputation as in Che et al, 2018
    time=False,
    mask=True,
    delta=True,
    seed=SEED,
)
train_loader = DataLoader(
    physionet2012, collate_fn=packed_sequence, batch_size=BATCH_SIZE
)
\end{minted}

Note that random seeds are set in the data class (line 23) to ensure reproducibility of the training and validation data sets and in PyTorch (line 15) for reproducible model training.

GRU-Simple is used in a binary classification context by passing the final hidden state of the GRU to a fully connected linear layer with sigmoid activation, as the aim is to train a model to predict the probability of in-hospital death.

\begin{minted}[linenos,firstnumber=28]{python}
class GRUSimpleBinaryClassifier(nn.Module):
    def __init__(self, input_size, hidden_size):
        super(GRUSimpleBinaryClassifier, self).__init__()
        self.gru = nn.GRU(
            input_size=input_size,
            hidden_size=hidden_size,
            batch_first=True,
        )
        self.linear = nn.Linear(in_features=hidden_size, out_features=1)

    def forward(self, x, hx=None):
        _, h_n = self.gru(x, hx)
        return torch.sigmoid(self.linear(h_n).squeeze(0))
\end{minted}

The PhysioNet 2012 challenge data has 45 channels.\footnote{One time channel, 37 time series channels and seven fixed variables: age, gender, height and type of ICU unit (one-hot encoded giving four channels).} For the GRU-Simple model, an observational mask and time delta is added for each channel resulting in $45 \times 3 = 135$ input channels.

\begin{minted}[linenos,firstnumber=40]{python}
model = GRUSimpleBinaryClassifier(
    input_size=135,
    hidden_size=HIDDEN_SIZE,
).to(device)
\end{minted}

The model is trained using the Adam optimiser and binary cross-entropy loss for 250 epochs. Predictions are made on the training data after each epoch and the area under the ROC curve is recorded.

\begin{minted}[linenos,firstnumber=44]{python}
optimizer = torch.optim.Adam(model.parameters(), lr=LEARNING_RATE)
loss_function = nn.BCELoss()

auc = np.full(N_EPOCHS, np.nan)
for epoch in range(N_EPOCHS):
    # Train model in batches
    for batch in train_loader:
        X = batch["X"].to(device)
        y = batch["y"].data.unsqueeze(1).to(device)
        optimizer.zero_grad()
        pred = model(X)
        loss = loss_function(pred, y)
        loss.backward()
        optimizer.step()
    model.eval()
    # AUC on validation data
    with torch.no_grad():
        X_val = physionet2012.X_val.to(device)
        pred_val = model(X_val)
        auc_epoch = roc_auc_score(physionet2012.y_val, pred_val.cpu())
        auc[epoch] = auc_epoch
    model.train()
print("Best AUC: {:.4f}".format(max(auc)))
\end{minted}

No attempt was made to tune hyper-parameters or optimise performance in this simple example, however the best model during training returns an AUC of 79.5\% on the validation data. The performance over five randomly generated seeds\footnote{293120, 339980, 555902, 711053 and 977556. You may not be able to replicate these results due to hardware differences.} is (80.4 $\pm$ 1.2)\%. This performance is consistent with the (80.8 $\pm$ 1.1)\% reported in \citet{horn2020} despite different data splits, hyper-parameters and other model design choices.

\section{Background information}
\label{background}

\torchtime{} is available from the Python Package Index. Installation instructions and documentation can be found at the project website, \url{https://philipdarke.com/torchtime}.

\subsection{Supported data sets}

All data sets in the UEA \& UCR time series classification repository \cite{Bagnall} and two PhysioNet repository \cite{Goldberger2000} data sets are supported:
\begin{itemize}
  \item Predicting Mortality of ICU Patients: The PhysioNet/Computing in Cardiology Challenge 2012\footnote{\url{https://physionet.org/content/challenge-2012/1.0.0/}} \cite{Silva2012}
  \item Early Prediction of Sepsis from Clinical Data: The PhysioNet/Computing in Cardiology Challenge 2019\footnote{\url{https://physionet.org/content/challenge-2019/1.0.0/}} \cite{Reyna2020}
\end{itemize}
In addition, a binary prediction variant of the PhysioNet 2019 challenge is provided. As in \citet{Kidger2020}, the first 72 hours of data are used to predict whether the patient develops sepsis at any point during hospitalisation.\footnote{Note that \citet{Kidger2020} appear to take the first 72 \textit{rows} of data rather than 72 hours as in \torchtime{}.}

\subsection{Batch first convention}

\torchtime{} uses the ``batch first'' convention under which data tensors are of shape $(n, s, c)$ where $n$ is number of batches, $s$ sequence length, and $c$ number of channels. Recurrent Neural Network models in PyTorch typically have a \texttt{batch\_first} argument which should be set to \texttt{True} when using \texttt{torchtime} data sets.

\subsection{Training and validation splits}

All UEA \& UCR repository data are provided in separate training and validation files in varying proportions. PhysioNet challenges are provided in two/three equal sized sets. \torchtime{} combines all data and takes samples to create training, validation and, if applicable, test data sets in the proportions specified by the user.

\subsection{Reproducibility}

Stratified sampling and missing data simulation are non-deterministic, however data sets are reproduced for a given random seed and care is taken to maintain this behaviour across releases. For completely reproducible results, all other non-deterministic behaviour should be controlled including batch generation and parameter initialisation. See \url{https://pytorch.org/docs/stable/notes/randomness.html} for guidance.

\section{Using \torchtime{}}
\label{using_torchtime}

Each data set in the \texttt{torchtime.data} module has a consistent API with attributes \texttt{X}, \texttt{y} and \texttt{length}:
\begin{itemize}
  \item \texttt{X} are the time series data in a tensor of shape $(n, s, c)$ where $n$ is the number of sequences, $s$ the (longest) sequence length and $c$ the number of channels. By default, the first channel is a time stamp and subsequent channels are as provided in the source data.\footnote{See appendix \ref{physionet2012} for PhysioNet 2012 challenge channel order.}
  \item \texttt{y} are target labels. These typically have shape $(n, l)$ where $l$ is the number of classes.\footnote{An exception is the PhysioNet 2019 challenge where a binary target is provided at each time point i.e. a tensor of shape $(n, s)$.}
  \item \texttt{length} are the length of each sequence in a tensor of shape $(n$).
\end{itemize}

\subsection{Creating a data set}

There are two required arguments, \texttt{split} and \texttt{train\_prop}. UEA \& UCL repository data sets also require the \argument{dataset} argument. See \url{https://www.timeseriesclassification.com/dataset.php} for a list of data sets.

The \argument{split} argument determines whether training, validation or, if applicable, test data are returned by the \texttt{X}, \texttt{y} and \texttt{length} attributes. The primary use of \texttt{split} is to specify the data returned when using a DataLoader (see \ref{using_dataloaders}).

Data splits can also be accessed explicitly by appending \texttt{\_train}, \texttt{\_val} or \texttt{\_test} to the attribute. For example, \texttt{X\_train} returns training time series and \texttt{y\_val} returns validation labels regardless of \texttt{split}.

Training and validation data sets are created by default. The \argument{train\_prop} argument sets the proportion of data allocated to training (see example \ref{ex:split1}). To create a training/validation/test split, also specify the proportion of data in the validation set using the \argument{val\_prop} argument (see example \ref{ex:split2}). Data splits are formed using stratified sampling.

\begin{listing}[h]
  \caption{PhysioNet 2012 challenge data with a 70/30\% training/validation split. The \texttt{val\_prop} argument is not required.}
  \label{ex:split1}
  \begin{minted}[linenos]{python}
from torchtime.data import PhysioNet2012

physionet2012 = PhysioNet2012(
    split="train",
    train_prop=0.7,
)
  \end{minted}
\end{listing}

\begin{listing}
  \caption{The UEA \& UCR repository data set ArrowHead with a 70/20/10\% training/validation/test split (note both \texttt{train\_prop} and \texttt{val\_prop} must be specified). ArrowHead is a univariate time series presented as a supervised classification problem with three classes. There are 148 sequences in the training data, each with 251 observations and two data channels (a time stamp/index followed by the time series).}
  \label{ex:split2}
  \begin{minted}[linenos]{python}
from torchtime.data import UEA

arrowhead = UEA(
    dataset="ArrowHead",
    split="train",
    train_prop=0.7,
    val_prop=0.2,
    seed=123,  # to reproduce example
)
  \end{minted}
  Accessing data with the \texttt{X}, \texttt{y} and \texttt{length} attributes:
  \begin{minted}{python}
>>> arrowhead.X  # shape (148, 251, 2)
tensor([[[  0.0000,  -1.8010],
         [  1.0000,  -1.7989],
         [  2.0000,  -1.7784],
         ...,
         [248.0000,  -1.7965],
         [249.0000,  -1.7985],
         [250.0000,  -1.8010]],
        ...,
        [[  0.0000,  -2.4343],
         [  1.0000,  -2.4315],
         [  2.0000,  -2.3426],
         ...,
         [248.0000,  -2.4596],
         [249.0000,  -2.4922],
         [250.0000,  -2.4644]]])

>>> arrowhead.y  # shape (148, 3)
tensor([[0., 0., 1.],
        ...
        [0., 0., 1.]])

>>> arrowhead.length  # shape (148)
tensor([251, ... 251])
  \end{minted}
\end{listing}

\subsection{Using DataLoaders}
\label{using_dataloaders}

Data sets are typically passed to a PyTorch \href{https://pytorch.org/docs/stable/data.html#torch.utils.data.DataLoader}{DataLoader} for model training. \texttt{torchtime.data} classes return batches as a dictionary of tensors \texttt{X}, \texttt{y} and \texttt{length}. The \texttt{split} argument determines whether training, validation or, if applicable, test data are returned when using a DataLoader.

It is recommended to use one instance of a \texttt{torchtime.data} class for a data set. If DataLoaders are required for validation and/or test splits, an efficient approach is to pass the validation/test data to \href{https://pytorch.org/docs/stable/data.html#torch.utils.data.TensorDataset}{TensorDataset} as in example \ref{ex:dataloaders}. This avoids holding multiple complete copies of the data set in memory.

\begin{listing}
  \caption{An efficient strategy to generate iterable DataLoaders for training, validation and test data splits. Note that \texttt{train\_dataloader} returns batches as a named dictionary, but \texttt{val\_dataloader} and \texttt{test\_dataloader} return a list \texttt{[X, y, length]}. Alternatively, the full training/test data can be accessed using the \texttt{X/y/length\_val} and \texttt{X/y/length\_test} attributes.}
  \label{ex:dataloaders}
  \begin{minted}[linenos]{python}
from torchtime.data import UEA
from torch.utils.data import DataLoader, TensorDataset

# Data set
arrowhead = UEA(
    dataset="ArrowHead",
    split="train",
    train_prop=0.7,
    val_prop=0.2,
    seed=123,  # to reproduce example
)

# Training data (see line 6)
train_dataloader = DataLoader(arrowhead, batch_size=32)

# Validation data
val_data = TensorDataset(
    arrowhead.X_val,
    arrowhead.y_val,
    arrowhead.length_val,
)
val_dataloader = DataLoader(val_data, batch_size=32)

# Test data
test_data = TensorDataset(
    arrowhead.X_test,
    arrowhead.y_test,
    arrowhead.length_test,
)
test_dataloader = DataLoader(test_data, batch_size=32)
  \end{minted}

Using the DataLoaders to iterate through batches:

  \begin{minted}{python}
>>> train_batch = next(iter(train_dataloader))
>>> train_batch["X"].shape
torch.Size([32, 251, 2])  # sequences are 251 long with 2 channels

>>> val_batch = next(iter(val_dataloader))
>>> val_batch[1].shape
torch.Size([32, 3])  # labels have 3 classes

>>> test_batch = next(iter(test_dataloader))
>>> test_batch[2].shape
torch.Size([32])  # length of each sequence
  \end{minted}

Alternatively, to access full validation/test data:

  \begin{minted}{python}
>>> arrowhead.X_val.shape
torch.Size([42, 251, 2])  # 42 sequences in validation data

>>> arrowhead.y_test.shape
torch.Size([21, 3])  # 21 sequences in test data
  \end{minted}
\end{listing}

\newpage
\subsection{Other options}

See Appendix \ref{api} for the full API. In summary:
\begin{itemize}
  \item Arguments are provided to impute and, for UEA \& UCR data sets, simulate missing data. See \ref{imputation} and \ref{simulate_missing} for more information.
  \item By default, a time stamp is appended to time series data as the first channel. This can be removed by setting the \argument{time} argument to \texttt{False}. Additional arguments are provided to add missing data masks and time delta channels, see \ref{masks}.
  \item The boolean argument \argument{standardise} standardises the time series. The training data mean $\mu^t$ and standard deviation $\sigma^t$ are calculated for each channel $c$ and the transform $(x_c - \mu_c^t) / \sigma_c^t$ applied to the training, validation and, if applicable, test data. By default, data is not standardised.
  \item Data splits are formed by stratified sampling. The \argument{seed} argument can be set to return reproducible splits. By default, no seed is set.
  \item When a data set is first initialised, data are downloaded, processed and cached in the \texttt{.torchtime} directory. By default the cache is placed the project root but this can be changed using the \argument{path} argument, for example to share a cache across projects. Set the \argument{overwrite\_cache} argument to \texttt{True} to refresh the cache.
\end{itemize}

\section{Working with missing data}
\label{missing_data}

\subsection{Observational masks and time delta channels}
\label{masks}

Patterns of missingness can be informative in some applications. For example, a doctor may be more likely to order a particular diagnostic test if they believe a patient has a medical condition. The presence or absence of this test result provides information about patient health in addition to its value.

Missing data/observational masks can be used to inform models of missing data. These are appended by setting the \argument{mask} argument to \texttt{True}.

\begin{listing}[H]
  \caption{Simulated missing data for the UEA \& UCR repository data set CharacterTrajectories (see \ref{simulate_missing}). The final three channels are an observational mask where 1.0 indicates data were recorded.}
  \label{ex:mask}
  \begin{minted}[linenos]{python}
from torchtime.data import UEA

char_traj = UEA(
    dataset="CharacterTrajectories",
    split="train",
    train_prop=0.7,
    missing=[0.8, 0.2, 0.5],  # simulate missing data
    mask=True,  # add observational mask channels
    seed=456,  # to reproduce example
)
  \end{minted}
  Example sequence (first five observations):
  \begin{minted}{python}
>>> char_traj.X[0, 0:5]
tensor([[ 0.0000,     nan,  0.1640,  0.6631,  0.0000,  1.0000,  1.0000],
        [ 1.0000, -0.0678,  0.2123,     nan,  1.0000,  1.0000,  0.0000],
        [ 2.0000, -0.1190,  0.2448,     nan,  1.0000,  1.0000,  0.0000],
        [ 3.0000,     nan,     nan,  1.0139,  0.0000,  0.0000,  1.0000],
        [ 4.0000,     nan,  0.2550,     nan,  0.0000,  1.0000,  0.0000]])
  \end{minted}
\end{listing}

Some models require the time since the previous observation (the time delta $\delta$), for example GRU-D \cite{Che2018}. This is added using the \argument{delta} argument. See \citet{Che2018} for implementation details.

The \texttt{time}, \texttt{mask} and \texttt{delta} arguments can be combined as required. The channel order is always time stamp, time series data, missing data mask then time delta.

\begin{listing}[H]
  \caption{The final three channels are time deltas $\delta_{t,c}$ for the \texttt{UEA} class in example \ref{ex:mask} with \texttt{delta=True} (rather than \texttt{mask}). The second time series channel is observed at times 2 and 4 therefore $\delta_{4,1}$ is 2 i.e. two time units since last observation. Note that $\delta_{0,c}=0~\forall~c \in X$ by definition.}
  \begin{minted}{python}
>>> char_traj.X[0, 0:5]
tensor([[ 0.0000,     nan,  0.1640,  0.6631,  0.0000,  0.0000,  0.0000],
        [ 1.0000, -0.0678,  0.2123,     nan,  1.0000,  1.0000,  1.0000],
        [ 2.0000, -0.1190,  0.2448,     nan,  1.0000,  1.0000,  2.0000],
        [ 3.0000,     nan,     nan,  1.0139,  1.0000,  1.0000,  3.0000],
        [ 4.0000,     nan,  0.2550,     nan,  2.0000,  2.0000,  1.0000]])
  \end{minted}
\end{listing}

\subsection{Imputation}
\label{imputation}

Off-the-shelf deep learning models are unable to handle missing values. A simple strategy to overcome this limitation is to impute missing values, and \torchtime{} supports ``zero'', mean, forward and custom imputation functions using the \argument{impute} argument. Imputation has no impact on the observational mask or time delta channels.

\begin{itemize}
  \item Under \textit{zero} imputation, missing data are replaced with the value zero.
  \item Under \textit{mean} imputation, missing data are replaced with the training data channel mean.
  \item Under \textit{forward} imputation, missing values are replaced with the previous channel observation. Note that this approach does not impute initial missing values, therefore these are replaced with the corresponding channel mean in the training data.
  \item Alternatively a \textit{custom imputation} function can be passed to impute. This must accept the arguments \texttt{X} (time series), \texttt{y} (labels), \texttt{fill} (the training data means/modes for each channel, see \ref{categorical}) and \texttt{select} (the channels to impute). It must return tensors \texttt{X} and \texttt{y} post imputation.
\end{itemize}

The mean and forward imputation implementations ensure that knowledge of the time series at times $t > i$ is not used when imputing values at time $i$. This is required when developing models that make online predictions.

\subsubsection{Handling categorical variables}
\label{categorical}

Mean imputation is unsuitable for categorical variables. The channel indices of categorical variables should be passed to the \argument{categorical} argument to impute values with the training data channel mode rather than the mean. This is also required for forward imputation to appropriately impute initial missing values. The calculated channel mean/mode can be overridden using the \argument{channel\_means} argument, for example to impute missing data with a fixed value.

\subsection{Simulating missing data for UEA \& UCR repository data sets}
\label{simulate_missing}

Data sets in the UEA \& UCR repository are typically regularly sampled and fully observed. To aid model development, missing data can be simulated in these data using the \argument{missing} argument. Data are dropped at random and replaced with \texttt{NaN}s.

\subsubsection{Regularly sampled data with missing time points}

If missing is a single value, data are dropped across all channels. This simulates regularly sampled data where some time points are not recorded for example dropped data over a network.

\begin{listing}[H]
  \caption{Simulated missing data for the UEA \& UCR repository data set CharacterTrajectories. Note that each time point is either fully observed or missing.}
  \label{ex:missing1}
  \begin{minted}[linenos]{python}
from torchtime.data import UEA

char_traj = UEA(
    dataset="CharacterTrajectories",
    split="train",
    train_prop=0.7,
    missing=0.5,  # 50% missing
    seed=123,  # to reproduce example
)
dataloader = DataLoader(char_traj, batch_size=32)
  \end{minted}
  Example sequence (first five observations):
  \begin{minted}{python}
>>> char_traj.X[0, 0:5]
tensor([[ 0.0000, -0.1849,  0.1978,  0.3263],
        [ 1.0000,     nan,     nan,     nan],
        [ 2.0000, -0.3744,  0.2511,  0.4260],
        [ 3.0000,     nan,     nan,     nan],
        [ 4.0000,     nan,     nan,     nan]])
  \end{minted}
\end{listing}

\subsubsection{Regularly sampled data with partial observation}

Alternatively, data can be dropped independently for each channel by passing a list representing the proportion missing for each channel. This simulates regularly sampled data with partial observation i.e. not all channels are recorded at each time point.

\begin{listing}[H]
  \caption{Simulated missing data for the \texttt{UEA} class in example \ref{ex:missing1} with \texttt{missing=[0.8, 0.2, 0.5]}. Note that each time point has a varying number of observations.}
  \begin{minted}{python}
>>> char_traj.X[0, 0:5]
tensor([[ 0.0000,     nan,  0.1978,  0.3263],
        [ 1.0000,     nan,  0.2399,     nan],
        [ 2.0000,     nan,  0.2511,     nan],
        [ 3.0000,     nan,     nan,  0.4016],
        [ 4.0000,     nan,     nan,  0.3410]])
  \end{minted}
\end{listing}

\section{Sequences of unequal length}
\label{unequal_length}

PhysioNet and some\footnote{For example CharacterTrajectories.} UEA \& UCR data sets feature sequences of unequal length. Tensors must be of regular shape, therefore sequences are padded\footnote{Padding is carried out at data set level to reduce the data processing required at batch generation. This increases memory usage if $\max(s_b) \ll \max(s_d)$ where $s_b$ and $s_d$ are sequence lengths for the batch and data set respectively.} to the length of the longest sequence with \texttt{NaN}s. The length of each sequence before padding is available using the \texttt{length} attribute.

Data sets of variable length can be efficiently represented in PyTorch using a \href{https://pytorch.org/docs/stable/generated/torch.nn.utils.rnn.PackedSequence.html}{PackedSequence} object. These are formed using \texttt{torch.nn.utils.rnn.pack\_padded\_sequence()} which by default expects the input batch to be sorted in descending length. Two collate functions are provided to support the use of PackedSequence objects in models:

\begin{itemize}
  \item \texttt{torchtime.collate.sort\_by\_length()} sorts each batch by descending length.
  \item \texttt{torchtime.collate.packed\_sequence()} returns \texttt{X} and \texttt{y} as a PackedSequence object.
\end{itemize}

Custom collate functions should be passed to the \texttt{collate\_fn} argument of a DataLoader.

\section{Conclusion}

\torchtime{} is a Python package providing PhysioNet challenge and UEA \& UCR repository data sets for use in PyTorch. Flexible data splits, missing data imputation, observational masks and time delta channels are supported. In addition, missing data can be simulated for UEA \& UCR repository data sets. \torchtime{} is well documented and open source under the MIT license.

Our aim is to support model development and benchmarking for complex time series data with a focus on Electronic Health Records. The longer term intention is to provide PyTorch implementations of relevant deep learning models. Feedback and suggestions for additional data sets, features or models are welcome.

\begin{ack}
  \torchtime{} uses some of the data processing ideas in \citet{Kidger2020} and \citet{Che2018}.

  This work is supported by the Engineering and Physical Sciences Research Council, Centre for Doctoral Training in Cloud Computing for Big Data, Newcastle University (grant number EP/L015358/1).
\end{ack}

\section*{Contributions}

Philip Darke conceptualised the work, developed \torchtime{}, drafted the manuscript and is the guarantor. Paolo Missier and Jaume Bacardit supervised the work. All authors critically revised the manuscript and approved the final version.

\bibliography{refs}
\bibliographystyle{unsrtnat}

\appendix
\newpage

\section{Implementation}

\subsection{Package structure}

The \texttt{torchtime.data} module contains the following classes that inherit from the PyTorch \texttt{torch.utils.data.Dataset} abstract class and can be passed to a DataLoader:
\begin{itemize}
  \item \texttt{torchtime.data.PhysioNet2012}
  \item \texttt{torchtime.data.PhysioNet2019}
  \item \texttt{torchtime.data.PhysioNet2019Binary}
  \item \texttt{torchtime.data.UEA}
\end{itemize}

The \texttt{torchtime.collate} module provides collate functions for working with sequences of unequal lengths and the \texttt{torchtime.impute} module provides helper functions for data imputation. Full API documentation is available at \url{https://philipdarke.com/torchtime}.

\subsection{High-level approach}

When creating an instance of a data class:

\begin{enumerate}
  \item \torchtime{} checks if the data set has been cached and, if so, loads and validates it by checking SHA256 checksums.
  \item If no cache is available, the checksum test fails or the argument \texttt{overwrite\_data} is set, data is downloaded, processed and combined to create a master data set. The tensors \texttt{X}, \texttt{y} and \texttt{length} as defined in section \ref{using_torchtime} are cached in the \texttt{.torchtime} directory.
  \item Missing data is simulated if applicable.
  \item Time stamp, observational mask and time delta channels are appended as required.
  \item Training, validation and test data splits are formed using stratified sampling.
  \item Data is standardised if specified.
  \item Missing data is imputed if specified.
  \item The attributes \texttt{X}, \texttt{y} and \texttt{length} are set to the data split specified by the \texttt{split} argument.
\end{enumerate}

Data format varies across PhysioNet challenges but EHR data are typically provided as a separate text file for each participant. \torchtime{} downloads all data from PhysioNet and iterates through each file to extract the time series data. The 2012 challenge requires additional processing as the data is in a ``long'' format, channels are not provided in order (see appendix \ref{physionet2012}) and the indicator $-1$ is used for missing data.

For UEA \& UCR repository data, a similar approach to \texttt{sktime} \cite{Loning2019} is used to download and extract the time series from the repository.

\subsection{Defining a new data set}

A data class is defined by inheriting the private class \texttt{torchtime.data.\_TimeSeriesDataset} and overloading the \texttt{get\_data} method. \texttt{get\_data} must download and process the data set and return the tensors \texttt{X}, \texttt{y} and \texttt{length} for the full data. The \texttt{torchtime.utils} module contains private helper functions to simplify the task.

\subsection{Development and quality control}

The primary dependencies are PyTorch \cite{Paszke2019}, \texttt{sktime} \cite{Loning2019} for extracting time series from \texttt{.ts} files and \texttt{scikit-learn} \cite{Pedregosa2011} for stratified sampling. Python Poetry was used for packaging.

To validate data integrity, the SHA256 is checked when a data set is loaded from cache. This prevents the accidental use of an invalid data set. To aid code quality, a comprehensive suite of unit tests are run to test core functionality and ensure reproducibility of data sets when making changes to the code base.

\torchtime{} is open source software under the MIT license and code is available at \url{https://github.com/philipdarke/torchtime}.

\subsection{Results}

The experiments in this manuscript were carried out on a machine running Ubuntu 22.04.4 LTS with a single NVIDIA RTX 3070 GPU and using \torchtime{} v0.5.0.

\section{API}
\label{api}

\subsection{Required arguments}

The\marginnote{dataset} data set to return. \textit{String.} \textbf{UEA \& UCR data sets only}.

The\marginnote{split} data split returned by attributes \texttt{X}, \texttt{y} and \texttt{length}. \textit{String: ``train'', ``val'' (validation) or ''test''.}

The\marginnote{train\_prop} proportion of data in the training set. \textit{Float.}

\subsection{Optional arguments}

The\marginnote{val\_prop} proportion of data in the validation set. \textit{Float.}

The\marginnote{missing} proportion of data to drop at random (see \ref{simulate_missing}). If \texttt{missing} is a single value, data are dropped from all channels. To drop data independently from each channel, pass the proportion missing for each channel in a list. \textit{Float or list of floats (default 0.0).} \textbf{UEA \& UCR data sets only}.

The\marginnote{impute} method used to impute missing data (see \ref{imputation}). \textit{Function or string: ``none'', ``zero'', ``mean'' or ``forward'' (default ``none'').}

Channel\marginnote{categorical} indices of categorical variables. Only required if imputing data. \textit{List of integers.} \textbf{UEA \& UCR data sets only}.

Override\marginnote{channel\_means} the calculated channel mean/mode, for example \texttt{\{1:4.5, 3:7.2\}} overrides channels 1 and 3 with the values 4.5 and 7.2 respectively. Only used if imputing data. \textit{Dictionary with integer keys and values (default \texttt{\{\}}).} \textbf{UEA \& UCR data sets only}.

Append\marginnote{time} a time stamp as the first channel. \textit{Boolean (default True).}

Append\marginnote{mask} an observational mask for each channel. \textit{Boolean (default False).}

Append\marginnote{delta} the time since previous observation for each channel. \textit{Boolean (default False).}

Standarise\marginnote{standardise} the time series. \textit{Boolean (default False).}

Refresh\marginnote{overwrite\_cache} the cached dataset. \textit{Boolean (default False).}

Path\marginnote{path} to the \texttt{.torchtime} cache directory. \textit{String (default ``.'' i.e. the project root directory).}

Random \argument{seed} for reproducibility. \textit{Integer (default ``none'').}

\section{PhysioNet 2012}
\label{physionet2012}

\subsection{Time series channels}

Data channels are in the following order:

\begin{description}[leftmargin=!,labelwidth=\widthof{\bf 22. NIDiasABP }]
  \item[~0. Mins] Minutes since ICU admission. Derived from the PhysioNet time stamp.
  \item[~1. Albumin] Albumin (g/dL).
  \item[~2. ALP] Alkaline phosphatase (IU/L).
  \item[~3. ALT] Alanine transaminase (IU/L).
  \item[~4. AST] Aspartate transaminase (IU/L).
  \item[~5. Bilirubin] Bilirubin (mg/dL).
  \item[~6. BUN] Blood urea nitrogen (mg/dL).
  \item[~7. Cholesterol] Cholesterol (mg/dL).
  \item[~8. Creatinine] Serum creatinine (mg/dL).
  \item[~9. DiasABP] Invasive diastolic arterial blood pressure (mmHg).
  \item[10. FiO2] Fractional inspired O$_2$ (0-1).
  \item[11. GCS] Glasgow Coma Score (3-15).
  \item[12. Glucose] Serum glucose (mg/dL).
  \item[13. HCO3] Serum bicarbonate (mmol/L).
  \item[14. HCT] Hematocrit (\%).
  \item[15. HR] Heart rate (bpm).
  \item[16. K] Serum potassium (mEq/L).
  \item[17. Lactate] Lactate (mmol/L).
  \item[18. Mg] Serum magnesium (mmol/L).
  \item[19. MAP] Invasive mean arterial blood pressure (mmHg).
  \item[20. MechVent] Mechanical ventilation respiration (0:false, or 1:true).
  \item[21. Na] Serum sodium (mEq/L).
  \item[22. NIDiasABP] Non-invasive diastolic arterial blood pressure (mmHg).
  \item[23. NIMAP] Non-invasive mean arterial blood pressure (mmHg).
  \item[24. NISysABP] Non-invasive systolic arterial blood pressure (mmHg).
  \item[25. PaCO2] Partial pressure of arterial CO$_2$ (mmHg).
  \item[26. PaO2] Partial pressure of arterial O$_2$ (mmHg).
  \item[27. pH] Arterial pH (0-14).
  \item[28. Platelets] Platelets (cells/nL).
  \item[29. RespRate] Respiration rate (bpm).
  \item[30. SaO2] O$_2$ saturation in hemoglobin (\%).
  \item[31. SysABP] Invasive systolic arterial blood pressure (mmHg).
  \item[32. Temp] Temperature (°C).
  \item[33. TroponinI] Troponin-I ($\mu$g/L). Note this is labelled \textit{TropI} in the PhysioNet data dictionary.
  \item[34. TroponinT] Troponin-T ($\mu$g/L). Note this is labelled \textit{TropT} in the PhysioNet data dictionary.
  \item[35. Urine] Urine output (mL).
  \item[36. WBC] White blood cell count (cells/nL).
  \item[37. Weight] Weight (kg).
  \item[38. Age] Age (years) at ICU admission.
  \item[39. Gender] Gender (0: female, or 1: male).
  \item[40. Height] Height (cm) at ICU admission.
  \item[41. ICUType1] Type of ICU unit (1: Coronary Care Unit).
  \item[42. ICUType2] Type of ICU unit (2: Cardiac Surgery Recovery Unit).
  \item[43. ICUType3] Type of ICU unit (3: Medical ICU).
  \item[44. ICUType4] Type of ICU unit (4: Surgical ICU).
\end{description}

Channels 38 to 41 do not vary with time. Channels 11 (GCS) and 27 (pH) are assumed to be ordinal and are imputed using the same method as a continous variable. Variable 20 (MechVent) has value \texttt{NaN} (the majority of values) or 1. It is assumed that value 1 indicates mechanical ventilation and \texttt{NaN} indicates either missing data or no mechanical ventilation. Accordingly, the channel mode is assumed to be zero. Channels 41 to 44 are the one-hot encoded PhysioNet variable ICUType.

\subsection{Outcome}

The outcome is the \textbf{In-hospital death} field (0: survivor, or 1: died in hospital).

\end{document}